\documentclass[runningheads]{llncs}

\usepackage{graphicx}
\usepackage{amsmath}
\usepackage[hidelinks]{hyperref}
\hypersetup{bookmarksdepth=2}
\usepackage{xspace}

\usepackage[]{algpseudocode}
\usepackage{algorithm} 
\usepackage{algorithmicx}
 
\newcommand{\set}[1]{\{\:#1\:\}}

\usepackage{scalerel}
\newcommand\fat[1]{\ThisStyle{\ooalign{%
  \kern.46pt$\SavedStyle#1$\cr\kern.33pt$\SavedStyle#1$\cr%
  \kern.2pt$\SavedStyle#1$\cr$\SavedStyle#1$}}}
\def\fvee{\mathbin{\fat{\vee}}}

\def \AND {\mathbin{\fat{\wedge}}}
\newcommand{\suchthat}{~\mathbin{\fat{|}}~}

\newcommand{\Templates}{\mathcal{T}}
\usepackage{xcolor}

\def \Constraints {X}
\def \svps {sv(p_1,\dots,p_n)}
\def \Chronicle {\mathcal{C}}
\def \sv {\mathit{sv}}
\def \loc {\mathit{loc}}
\def \Go {\mathit{Go}}

\begin{document}
\title{A Constraint-based Encoding for Domain-Independent Temporal Planning}

\author{Arthur Bit-Monnot}
\authorrunning{A. Bit-Monnot}

\institute{
    University of Genoa, Genoa, Italy\\
    University of Sassari, Sassari, Italy\\
    \email{afbit@uniss.it}
}
\maketitle              
\begin{abstract}
We present a general constraint-based encoding for domain-independent task planning.
Task planning is characterized by causal relationships expressed as conditions and effects of optional actions. Possible actions are typically represented by templates, where each template can be instantiated into a number of primitive actions.

While most previous work for domain-independent task planning has focused on primitive actions in a state-oriented view, our encoding uses a fully lifted representation at the level of action templates. It follows a time-oriented view in the spirit of previous work in constraint-based scheduling.

As a result, the proposed encoding is simple and compact as it grows with the number of actions in a solution plan rather than the number of possible primitive actions. When solved with an SMT solver, we show that the proposed encoding is slightly more efficient than state-of-the-art methods on temporally constrained planning benchmarks while clearly outperforming other fully constraint-based approaches.
\end{abstract}

\section{Introduction} 

Task planning is a field of Artificial Intelligence concerned with finding a set of actions that would result in desirable state.
Its key difficulty lies in the handling of causal relationships between a large number of potential actions.
Research in the field has focused primarily on the definition of relaxations of the classical planning problem in order to define accurate heuristic functions to guide tree search algorithms.
Those heuristics have proved to be highly effective for reasoning on the causal relationships that occur in task planning and have driven most research to focus on state-based heuristic search.

Despite their success in classical planning, they have proved to be difficult to extend to more expressive models including time, resources or continuous changes.
This led to a renewal of interest into constraint-based models and search as a way to increase the expressiveness of domain-independent task planners \cite{cashmore2016,dvorak-2014-ictai,scala2016a}.

In this paper, we propose an encoding for the causal relationships that are at the core of domain-independent planning.
Unlike the most recent work proposing compilations of domain-independent temporal planning to Constraint Satisfaction Problems (CSPs) which follow a state-oriented view \cite{bryce2015a,cashmore2016}, we use a time-oriented view inspired by the work in the field of constraint-based planning and scheduling \cite{cesta2009,frank2003,ghallab1994,laborie2008}.
The key benefit of this encoding is its simplicity and its good fit for integration into more general constraint satisfaction problems, going one step closer to bridging the gap between planning and scheduling.

We start by giving a background on task planning and the representation of temporal planning problems as chronicles.
We then show how such planning problems can be concisely encoded into CSPs, using a fully lifted representation based on chronicles.
The resulting encoding is leveraged in a domain-independent planner for ANML, an expressive language for the specification of planning problems \cite{anml2008}.
Using an off-the-shelf SMT solver as a backend, the planner is shown to outperform state-of-the-art temporal planners on temporally constrained planning benchmarks.
Last, we discuss the related work in constraint-based planning and scheduling, especially analyzing other domain-independent planners that rely on SMT.

\section{Background}

\subsection{A Distilled Planning Problem}

\def \States {S}

In its simplest formulation, a planning problem is composed of 
\emph{(i)} an initial state, 
\emph{(ii)} a goal state and 
\emph{(iii)} a set of primitive actions.

A state is an assignment to a set of \emph{state variables}, each denoting one particular feature of the environment (e.g. the location of a truck).
We denote as $\States$ the set of possible states.

A primitive action $a$ is composed of:
\begin{itemize}
    \item a set of conditions over state variables. It characterizes the set of states $\States_a^\mathit{app} \subseteq \States$ in which the action is applicable.
    \item a set of effects from which one can construct a state transition function $f_a : \States_a^\mathit{app} \rightarrow S$
\end{itemize} 

Given an initial state $s_0 \in \States$, a goal state $s_g \in \States$ and a set of primitive actions, a plan is a sequence of primitive actions $\langle a_1, \dots, a_n \rangle$.
For a given plan, one can build the resulting sequence of states $\langle s_1, \dots, s_n\rangle$ such that $s_i = f_{a_i}(s_{i-1})$.
A plan is a solution to a planning problem if all actions are applicable in the previous state ($\forall_{i \in [1,n]} s_{i-1} \in \States_{a_i}^\mathit{app}$) and the final state is the goal state ($s_n = s_g$).

This formulation of planning as a state transition system has been at the core of domain-independent planning research, and of PDDL, the de-facto standard language for describing planning problems \cite{mcdermott1998}.
Several extensions have been devised for extending the expressiveness of PDDL, most notably to handle a limited form of temporal constructs and numeric variables \cite{edelkamp2004,fox2003}.
Nevertheless, the focus of the language and the benchmarks of the International Planning Competition remains heavily focused on handling the causal relationships derived from the conditions and effects of primitive actions.

\subsection{Temporal Planning as Chronicles}

An other notable representation for planning follows a time-oriented view as opposed to the state-oriented view above.
We here detail the representation of planning problems as chronicles \cite{ghallab2004}, which was first introduced in the IxTeT planner \cite{ghallab1994}.
This representation avoids direct references to states and instead describes the evolution of the environment through temporally qualified assertions over the state variables.

A \emph{type} is defined by a set of values.
A type is either 
\emph{(i)} a set of domain constants (e.g. the type $\mathit{Truck} = \set{R_1, R_2}$ defines two truck objects $R_1$ and $R_2$ in the planning problem),
\emph{(ii)} a discrete or continuous set of numeric values, or 
\emph{(iii)} a representation of temporal instants, for which we typically consider the set of natural numbers.%
\footnote{Note that this integer based representation of time is no less expressive than a real-valued representation when forbidding instantaneous changes, as common in temporal planning \cite{Cushing2012}.}

A (decision) \emph{variable} is associated with a type that defines its initial domain.
We denote as \emph{timepoints} the subset of variables that represent a temporal instant.

A \emph{state variable} denotes the evolution of a particular state feature over time.
A state variable is typically parameterized by one or multiple domain objects to represent the state of a particular object in a planning problem.
For instance $\loc(R_1)$ denotes the evolution of the location of the truck $R_1$ over time.
A state variable expression can be parameterized by decision variables, in which case the actual state variable it refers to depends on the value taken by its parameters, e.g., $\loc(r)$ will refer to $\loc(R_1)$ or $loc(R_2)$ depending on the value taken by the variable $r$ of type $\mathit{Truck}$.

\subsubsection{Chronicle}

In line with previous work in temporal planning, we represent the core constructs of a planning problem with chronicles \cite{ghallab2004}.
A chronicle is a tuple $(V,\Constraints,C,E)$ where: 
\begin{itemize}
    \item $V$ is a set of variables.
    \item $\Constraints$ is a set of constraints over the variables in $V$.
    \item $C$ is a set of conditions. Each condition is of the form $[s, e]\ \sv(p_1,\dots,p_n) = v$ where $s$ and $e$ are timepoints in $V$, $\sv(p_1,\dots,p_n)$ is a parameterized state variable (with each $p_i \in V$) and $v \in V$ is a variable.
    A condition states that the state variable $\svps$ must have the value $v$ over the temporal interval $[s,e]$.
    \item $E$ is a set of effects. Each effect is of the form $[s, e]\ \sv(p_1,\dots,p_n) \gets v$ where $s$ and $e$ are timepoints, $\sv(p_1,\dots,p_n)$ is a parameterized state variable (with each $p_i \in V$) and $v \in V$ is a variable.
    Such an effect states that the state variable $\svps$ will take the value $v$ at time $e$.
    Over the temporal interval $]s,e[$ the state variable is transitioning from its previous value to $v$: it has an undefined value and cannot be further constrained.
    
\end{itemize}
A chronicle is thus a constraint satisfaction problem $(V,\Constraints)$ extended with additional constructs to represent the conditions and effects that are at the core of planning problems.%
\footnote{
    The original chronicle model used transitions instead of effects.
    We use effects to more closely match the classical definition of planning problems and simplify the presentation.
    Note that transitions can still be straightforwardly encoded by combining a condition and an effect.
}

\paragraph{Action Chronicle}
A planning problem defines a set of action templates $\Templates$ where each action template can be instantiated into chronicles.
We denote as $\Chronicle_a^i$ a chronicle instantiated from an action template $a \in \Templates$, where $i$ distinguishes chronicles instantiated from the same template.

Considering an action template $\Go \in \Templates$ that moves a truck $r$ from a location $\ell_s$ to a location $\ell_e$, its instantiation as a chronicle $\Chronicle_\Go$ could have the following components:
\begin{itemize}
    \item $V = \set{ r, \ell_s, \ell_e, t_s, t_e }$ where $r, \ell_s, \ell_e$ are variables representing the parameters of the action (truck, start location and end location) and $t_s$ and $t_e$ are timepoints representing the start and end times of the action.
    \item $\Constraints = \set{ t_e = t_s +10, \ell_s \neq \ell_e }$, constraints stating that the action takes 10 time units and that the origin $\ell_s$ and destination $\ell_e$ must be different.
    \item $C = \set{ [t_s, t_s]\ \loc(r) = \ell_s }$, i.e., the truck $r$ must be in location $\ell_s$ at the action' start $t_s$.
    \item $E = \set{ [t_s, t_e]\ \loc(r) \leftarrow \ell_e }$, i.e., the truck $r$ will be in location $\ell_e$ at the action's end $t_e$. Its location is undefined for the duration of the action $]t_s,t_e[$.    
\end{itemize}
Such a chronicle is called an \emph{action chronicle}.

\paragraph{Initial Chronicle}
We distinguish action chronicles from the initial chronicle $\Chronicle_0$ that represents the initial state and the planning objectives.
In addition, the initial chronicle might provide a partial view of the expected state evolution, e.g., that a cargo ship (outside the control of the planner) will arrive at 5pm.

Specifying a problem where the truck $R_1$ is initially in location $L_0$ and must be in location $L_2$ or $L_3$ before time 100 is encoded with the chronicle $\Chronicle_0 = (V_0, \Constraints_0, C_0, E_0)$ where:
\begin{itemize}
    \item $V_0 = \set{ t,\ell }$
    \item $\Constraints_0 = \set{t < 100,\ \ell = L_2 \vee \ell = L_3 }$, constraints restricting the solution set.
    \item $C_0 = \set{ [t,t]\ \loc(R_1) = \ell }$, condition specifying the goal.
    \item $E_0 = \set{ [0,0]\ \loc(R_1) \leftarrow L_0}$, effect defining the initial state: the truck $R_1$ is in $L_0$ at the initial time.
\end{itemize}

A planning problem can be represented as a pair $(\Chronicle_0, \Templates)$ where $\Chronicle_0$ defines the initial state and objectives and $\Templates$ is a set of action templates that can be instantiated into action chronicles.

\section{Planning as a Constraint Satisfaction Problem}
\newcommand{\prez}{\mathit{o}}

One of the difficulty that arises in planning is that the number of actions is not known beforehand nor can it be given tight bounds.
As a first step, we consider a \emph{bounded} planning problem composed of an initial chronicle $\Chronicle_0$ and finite set of  \emph{optional} action chronicles $AC$.
An optional action chronicle $\Chronicle_a^i \in AC$ is a action chronicle associated to a boolean variable $\prez_a^i$ that is true ($\top$) if $\Chronicle_a^i$ is part of the solution plan and false ($\bot$) otherwise.

Consider the planning problem of moving the truck $R$ to either $L2$ or $L3$ from the previous section.
Considering a set of optional action chronicles $\set{ \Chronicle_\Go^1, \Chronicle_\Go^2 }$, we can represent all the plans composed of 0, 1 or 2 instances of the $\Go$ action.
The actual plan depends on the instantiation of the decision variables.
One could build a satisfying solution from the partial assignment $\set{ \prez_\Go^1 \gets \top,\ r^1 \gets R_1,\ \ell_s^1 \gets L_0,\ \ell_e^1 \gets L_2,\ \prez_\Go^2 \gets \bot}$ where $r^1$, $\ell_s^1$ and $\ell_e^2$ are the eponymous variables in $\Chronicle_\Go^1$. 
Such an assignment states that only the first action would be present and would move the truck $R_1$ from $L_0$ to $L_2$.
Note that this only partially defines the solution plan and other assignments would need to be made, notably to the action timepoints and to the variables of $\Chronicle_0$.

A \emph{bounded} planning problem $\Pi$ is the set of chronicles $\set{\Chronicle_0} \cup AC$ where each $\Chronicle \in \Pi$ is associated to a boolean variable $present(\Chronicle)$ that is: $\prez_a^i$ if $\Chronicle$ is an action chronicle $\Chronicle_a^i \in AC$; or $\top$ if $\Chronicle = \Chronicle_0$.
This planning problem is called \emph{bounded} as there is an upper limit on the number of actions in a solution plan defined by the number of action chronicles.

\subsection{Building Blocks}

\newcommand{\svp}[1]{\sv ( p_1#1, \dots, p_n#1 )}
\newcommand{\efftok}[1]{\langle \prez{#1} : [s{#1},e{#1},t{#1}]\ \svp{#1} \leftarrow v{#1} \rangle}
\newcommand{\condtok}[1]{\langle \prez{#1} : [s{#1},e{#1}]\ \svp{#1} = v{#1} \rangle}

Given a bounded planning problem  $\Pi$, we now describe the core structures for compiling it to a constraint satisfaction problem.

\subsubsection{Condition Token}
Given a chronicle $\Chronicle \in \Pi$, each condition $[s,e]\ \svp{} = v$ in $\Chronicle$ is associated to a \emph{condition token}:
\begin{equation*}
    present(\Chronicle): [s,e]\ \svp{} = v
\end{equation*}
This token represents that, if $\Chronicle$ is part of the solution ($present(\Chronicle) = \top$), then the state variable $\svp{}$ must have the value $v$ over the temporal interval $[s,e]$.

We denote as $C_\Pi$ the set of condition tokens in $\Pi$.

\subsubsection{Effect Token}
Given a chronicle $\Chronicle \in \Pi$, we associate to each effect $[s,e]\ \svp{} \gets v$ in $\Chronicle$ an \emph{effect token}:
\begin{equation*}
    present(\Chronicle) : [s,e,t]\ \svp{} \gets v
\end{equation*}
where $t$ is a new timepoint variable.
This token states that, if $\Chronicle$ is part of the solution ($present(\Chronicle) = \top$), then the state variable $\svp{}$ is undefined over the temporal interval $]s,e[$ and has the value $v$ over the temporal interval $[e,t]$.
The introduction of the new decision variable $t$ allows us to encode a minimal time for which the effect will persist.

We denote as $E_\Pi$ the set of effect tokens in $\Pi$.

\subsubsection{Characteristics of Tokens}
Effect tokens here represent the evolution of state variables over time.
Each (present) effect token imposes a new value to its state variable.
This value is constrained to be maintained on a given temporal interval.
Effect tokens thus encode the state evolution.
On the other hand, condition tokens place constraints on the state evolution by requesting a given state variable to have a given value over a temporal interval.
Intuitively, such a condition is achieved if there is a corresponding effect token that imposes the appropriate value.

For a given token, the variables involved allow four degrees of freedom on which a solver might play to build a consistent plan:
\begin{itemize}
    \item the presence of the token (variable $present(\cdot)$).
    \item the temporal interval on which it applies, (variables $s$, $e$ and $t$).
    \item the state variable on which it applies (variables $p_1,\dots,p_n$).
    \item the value taken by the state variable (variable $v$).
\end{itemize}
Those variables are typically shared with other tokens from the same action chronicle.
These interdependencies between tokens are at the core of the hardness of planning as having an effect token to establish a condition requires the presence of all other tokens from the same chronicle.
This in turn requires new conditions to be established as well as new effect tokens that might interact with the existing ones.

\subsection{Constraints for Plan Consistency}

Given a bounded planning problem $\Pi$, we define a set of constraints that encode the requirements for a plan to be consistent.

\subsubsection{Coherence Constraint}
State variables are similar to unary resources in that they can only take a single value at a given time.
Any two distinct effect tokens $\alpha$ and $\alpha'$ in $E_\Pi$ must be coherent: they may not concurrently impose a value to the same state variable.
\begin{align*}
    \text{Given }  \alpha &= \efftok{} \in E_\Pi\\
           \alpha' &= \efftok{'} \in E_\Pi
\end{align*}           
then the constraint $\textit{coherent}(\alpha,\alpha')$ is defined as:
\begin{equation*}
    \prez \AND \prez' ~\implies~ t \leq s' \fvee t' \leq s \fvee p_1 \neq p_1' \fvee \dots \fvee p_n \neq p_n'
\end{equation*}
The coherence constraint enforces that a given state variable has at most one value at any point in time.
It is done by forcing two effect tokens to be non overlapping which can be enforced along three axes: presence ($\prez$), time ($]s,t]$) and state variable ($\svp{}$).

\subsubsection{Support Constraint}
We say that a condition token $\beta \in C_\Pi$, is \emph{supported by} an effect token $\alpha' \in E_\Pi$ if $\alpha'$ establishes the value required by $\beta$ and that this value persists for the duration of $\beta$.
\begin{align*}
    \text{Given }  \beta &= \condtok{} \in C_\Pi \\
                 \alpha' &= \efftok{'} \in E_\Pi
\end{align*}
then $\textit{supported-by}(\beta,\alpha')$ is defined as: 
\begin{equation*}
    \prez' \AND e' \leq s \AND e \leq t' \AND p_1 = p_1' \AND \dots \AND p_n = p_n' \AND v = v'
\end{equation*}
Less formally, it means that $\beta$ is supported by $\alpha'$ if
\emph{(i)} $\alpha'$ is present,
\emph{(ii)} $\beta$ is contained in the interval $[e',t']$ over which the effect of $\alpha'$ persists, and
\emph{(iii)} $\alpha'$ establishes the right value on the right state variable.

The fact that a condition token $\beta \in C_\Pi$ must be supported by at least one effect token if it is present is encoded by the constraint $\textit{supported}(\beta)$:
\begin{equation*}
     \textit{present}(\beta) \implies \bigvee_{\alpha \in E_\Pi} \textit{supported-by}(\beta, \alpha)
\end{equation*} 

\subsubsection{Internal Chronicle Consistency}
Given an optional chronicle $\Chronicle = (V, X, C, E)$, if it is part of the solution then all its internal constraints must hold, which is represented by the constraint $\textit{consistent}(\Chronicle)$:
\begin{equation*}
    \textit{present}(\Chronicle) \implies \bigwedge_{c \in X} c
\end{equation*}

\subsubsection{Formulation as a CSP}

A bounded planning problem $\Pi$ is encoded as a Constraint Satisfaction Problem $(V_\Pi, X_\Pi)$ where:
\begin{align*}
    V_\Pi  = & ~\set{ V \suchthat (V,\Constraints,C,E) \in \Pi} \\
           \cup &~\set{ \textit{present}(\Chronicle) \suchthat \Chronicle \in \Pi } \\[5pt]
    \Constraints_\Pi = &~  \set{ \textit{coherent}(\alpha,\alpha') \suchthat \alpha,\alpha' \in E_\Pi, \alpha \neq \alpha'} \\
                     \cup &~ \set{ \textit{supported}(\beta) \suchthat \beta \in C_\Pi } \\
                     \cup &~ \set{ \textit{consistent}(\Chronicle) \suchthat \Chronicle \in \Pi }
\end{align*}
 
\subsection{Symmetry Breaking Constraints}

A given bounded planning problem $\Pi$ might contain several action chronicles of the same action template, e.g., several chronicles being instantiations of the $\Go$ action template.
In the current formulation, nothing distinguishes two action chronicles of the same template and any satisfying assignment can be made into a new one by exchanging the variables of the two chronicles.

Given the set of action templates $\Templates$, for any action template $a \in \Templates$, we refer to all action chronicles of the template $a$ in $\Pi$ as $\langle \Chronicle_a^1, \dots, \Chronicle_a^n \rangle$, in an arbitrary order.

From this ordering, we define two symmetry breaking constraints.
The first one requires that, if a given chronicle is present, then all its predecessors in the ordering are present as well:
\begin{equation*}
    \bigwedge_{a \in \Templates} \ \bigwedge_{i \in [1,n-1]} present(\Chronicle_a^{i+1}) \implies present(\Chronicle_a^i)
\end{equation*}

The second one requires the ordering on chronicles to match the ordering of actions in the solution plan:
\begin{equation*}
    \bigwedge_{a \in \Templates} \ \bigwedge_{i \in [1,n-1]} start(\Chronicle_a^{i+1}) \geq start(\Chronicle_a^i)
\end{equation*}
where $start(\Chronicle)$ is the start timepoint of the action chronicle $\Chronicle$.

\section{Instantiation in a Domain-Independent Planner}

We now describe how the presented encoding can be leveraged in a fully domain-independent temporal planner.
We present LCP (Lifted Constraint Planner), a planner that solves ANML planning problems using an SMT solver as a backend.
Rather than providing a fully fledged temporal planner, LCP aims at validating the effectiveness of our encoding on standard temporal planning benchmarks.

\subsection{The ANML Language}

The Action Notation Modeling Language (ANML) is a proposal by NASA Ames Research Center to represent rich temporal planning and scheduling problems \cite{anml2008}.
ANML features a clear notion of actions with conditions and effects integrated with a rich representation of time.

Compared to the temporal variants of PDDL, the ANML language has two important features.
The support of multi-valued states variables, as opposed to the boolean state variables of PDDL, allows for a more compact representation of the state.
In addition, ANML supports referencing timepoints other than the start and end of an action, thus increasing the capabilities to model complex temporal actions.
Temporal PDDL planning problems are usually easy to express in ANML, with some recent work to provide an automated translation \cite{Bernardini2018,Bernardini2011}. 

Translation of ANML planning problems into chronicles is straightforward as all constructs we are interested in have a one to one mapping \cite{bit-monnot2016}.%
\footnote{Omitted in our translations are the hierarchical and resource constructs of ANML that are beyond the scope of this paper.}
We parse an ANML problem file into the initial chronicle $\Chronicle_0$ and a set of action templates $\Templates$ that can be instantiated into action chronicles.

\subsection{Solving with SMT}

Our solving procedure works by incrementally generating bounded planning problems $\Pi_k$ with an increasing number of optional actions. For each problem, an SMT solver is used to either prove the consistency of $\Pi_k$ or the absence of a satisfying assignment.
If $\Pi_k$ is consistent, the solution plan is extracted from the found satisfying assignment.
Otherwise the process continues with the next bounded problem $\Pi_{k+1}$.

The overall procedure is given in Algorithm~\ref{alg:planner}.
The planner iteratively generates bounded planning problems of increasing depth $k$ until a solution is found or a depth $k_{\max}$ is reached.
Given an input planning problem $(\Chronicle_0,\Templates)$, we generate a bounded problem $\Pi_k$ using the procedure $\textsc{GenProblem}(\Chronicle_0, \Templates, k)$ as follows:
\begin{align*}
    \Pi_k = &~ \set{ \Chronicle_0} \cup  \bigcup_{a \in \Templates} \set{ \Chronicle_a^i \suchthat i \in [1,k]}
\end{align*}
For a problem $\Pi_k$, this formulation allows the presence of $k$ instances of each action template $a \in \Templates$.
The $\textsc{CheckSMT}(\Pi_k)$ function encodes $\Pi_k$ as a CSP using the encoding we presented.
The consistency of the resulting CSP is checked with the Z3 SMT solver \cite{demoura2008}.

\begin{algorithm}[htb!]
    \caption{
        Planning procedure of LCP (Lifted Constraint Planner) for an input problem $(\Chronicle_0,\Templates)$ and maximum depth $k_{\max}$.
    }
    \label{alg:planner}
    \begin{algorithmic}
        \Function{LCP}{$\Chronicle_0$, $\Templates$, $k_{\max}$}
        \State $k \gets 0$
        \While{$k \leq k_{\max}$}
            \State $\Pi_k \gets \Call{GenProblem}{\Chronicle_0, \Templates, k}$
            \State $\mathit{model} \gets \Call{CheckSMT}{\Pi_k}$
            \If{$\mathit{model}$ is a consistent model}
            \State \Return \Call{ExtractSolution}{$\mathit{model}$}
            \EndIf
            \State $k \gets k+1$
        \EndWhile 
        \State \Return failure
        \EndFunction
    \end{algorithmic}
\end{algorithm}

\subsection{Limitations}

As it stands, LCP is a rather na\"ive instantiation of the proposed encoding into a domain-independent planner.
One of the limitation is the absence of reuse of the inconsistent models of previous steps when generating a new bounded planning problem.
At the very least, one could analyze the unsatisfiable core provided by SMT solvers to infer which actions must be added to the next bounded planning problem.

More importantly, the use of SMT is certainly suboptimal given the progress made in constraint-based scheduling solvers in tackling closely related problems. Indeed, solvers such as CP Optimizer provide constructs for optional temporal intervals and state functions that closely relate to the building blocks of our encoding \cite{laborie2008,laborie2009}.
However, as we further discuss in \autoref{sec:related-work}, there is still a mismatch between the available global constraints and the needs for domain-independent planning.
In the absence of global constraints for CP solvers, SMT solvers provide a good backup solution for our encoding that features many disjunctive constraints.

LCP here simply aims at a preliminary evaluation of the proposed encoding against existing planners.

\section{Experiments}

\newcommand{\plus}{\texttt{+}\xspace}
\newcommand{\smtplan}{SMTPlan\plus}
\newcommand{\airports}{\textit{airport-timewindows}\xspace}
\newcommand{\satellites}{\textit{satellite-timewindows}\xspace}
\newcommand{\pipes}{\textit{pipesworld-deadlines}\xspace}

\subsection{Comparison with State of the Art Temporal Planners}

We evaluate our approach against several state of the art temporal planners on benchmarks from the International Planning Competition (IPC).
We focus on problems where time plays an important role in the solution: planning problems that have deadlines or time windows that constrain the occurrence time of actions in the plan.
The rationale for doing so is that planning problems without such features can be solved without any explicit handling of time (with the exception of some problem with required concurrency \cite{Cushing2007}). 

We compare ourselves to Temporal Fast Downward (TFD) \cite{Eyerich2012}, OPTIC \cite{Benton2012}, FAPE \cite{bit-monnot2016,dvorak-2014-ictai} and \smtplan \cite{cashmore2016}. 
TFD is a forward-search state-space planner that ranked second in the last temporal track of the IPC \cite{vallati2015}. Its search is based on the addition of new primitives actions at the end of an existing plan and heavily relies on heuristics for guidance.
OPTIC is similar in its use of forward-search but uses a different heuristic and a lifted temporal representation handled in an STN.
It is an improved version of POPF \cite{coles2010}, runner up in the temporal track of the penultimate IPC.

FAPE represents another line of temporal planners that uses a constraint-based representation.
FAPE plans in the space of lifted partial plans, where each plan is composed of a set of partially instantiated actions whose variables are embedded in a CSP.
Search proceeds by either imposing values on variables or extending the partial plan with additional actions, thus extending the CSP with new constraints, variables and values in previous domains.
FAPE is the first planner in this line of work to show good performance in a domain-independent setting \cite{bit-monnot2016}.

\smtplan \cite{cashmore2016} is a recent planner that supports the full range of PDDL\plus \cite{fox2006} through a compilation to SMT.
Unlike LCP, its encoding is state-oriented with additional constructs to support non-linear continuous change.
Given \smtplan's usage of SMT to solve a constraint-based encoding, an additional, more detailed, comparison will be made in the next subsection.

We use the \airports, \satellites and \pipes domains from IPC4 which feature durative actions and temporal constraints on the plan in the form of deadlines and time-windows in which actions can be performed.
The \airports domain focuses on planning the takeoff and landing of several planes in an airport, the movement of planes being temporally constrained by the arrival of other planes.
The \satellites domain plans observations and data transmission of a fleet of satellites subject to visibility windows.
The \pipes domain focuses on planning the transportation of products through pipes, subject to delivery deadlines.
TFD, OPTIC and \smtplan use the original PDDL domains while LCP and FAPE use their translation to ANML from FAPE's benchmark problems.%
\footnote{\url{https://github.com/laas/fape/tree/master/planning/domains}}

As standard in the IPC, we evaluate the planners on the number of problems solved with a 30 minutes timeout.
Benchmarks are run on an Intel i5-7200U @ 2.50GHz CPU with 8Gb of RAM. 
LCP and \smtplan use Z3 \cite{demoura2008} in version 4.6.3 as a black-box SMT solver.

Results are given in \autoref{tab:solved}.
It can be seen that, on temporally constrained problems, LCP is slightly ahead of both FAPE and OPTIC in terms of number of problems solved.
TFD performs poorly on such problems while \smtplan fails to solve any problem.

LCP performs best in the \pipes domain which is characterized by much interference between the available actions, reducing the ability to reason independently on goals.
Its worst performance is on the \satellites domain whose difficulty lies in the number of mostly independent goals requiring a large number of actions.
In this setting, LCP fails to prove the absence of solution before reaching the subproblem that actually contains one.
The \airports domain is an intermediate between those two domains making LCP performs only slightly better than other planners.

\begin{table}[htb]
    \centering
    \setlength{\tabcolsep}{8pt}
    \renewcommand{\arraystretch}{1.3}
    \begin{tabular}{|c|c|cccc|} \hline
                    & LCP         & FAPE        & OPTIC       & TFD   & \smtplan\\ \hline
     \airports      & \textbf{8}  & 7           & 7           & 1     & 0\\
     \satellites    & 4           & \textbf{10} & 4           & 0     & 0\\
     \pipes         & \textbf{17} & 6           & 13          & 2     & 0\\ \hline
     \textbf{Total} & \textbf{29} & 23          & 24          & 3     & 0\\ \hline
    \end{tabular}\\[5pt]
    \caption{
        Number of problems solved by the considered planners within 30 minutes.
        Best result is given in boldface.
    }
    \label{tab:solved}
\end{table}

\subsubsection{Note on expressiveness}
Unlike PDDL temporal planners, LCP supports rich temporal constructs, including actions with conditions and effects besides the start and end timepoints of actions.
It leverages ANML's multi-valued state variables (as opposed to the boolean state variables of PDDL) to provide a more compact representation.
Note that FAPE, which also uses the ANML language as input, has the same characteristics.

In addition, LCP readily supports numeric variables both as action parameters and state variables.
State of the art temporal planners do not support numeric parameters due to their need to ground the problem to derive heuristics, even though some recent work was done in this direction for forward-search planners \cite{savas2016}.

\subsubsection{Note on performance in non-temporal domains}
We also tested the performance of LCP on non-temporal domains, i.e., domains where time is either absent or limited to the duration of actions and can soundly be omitted \cite{Cushing2007}.
On the domains tested (\textit{blocks}, \textit{logistics} and \textit{rovers} from the IPC), LCP solved around two thirds of the problems solved by FAPE, OPTIC and TFD who all had similar performance.
As could be expected, LCP does not reach the performance of state of the art planners on non-temporal problems. 
It nevertheless showed consistent performance in its handling of causal relationships.
 
\subsection{Comparison with \smtplan}
\label{sec:comparison-smtplan}

Our work was motivated by the recent developments in the planning community to extend the expressiveness of task planners.
Research in domain-independent planning has been mostly focused on ground state-based heuristic search.
Such planners deliberately choose to cope with very large search spaces because it enables the definition of very effective heuristic functions to guide their exploration.

The will to support more realistic problems through extensions to PDDL is however problematic.
Indeed extensions affect the performance of heuristics that become less informative, with dramatic effects on search performance.
This state of affairs has motivated the introduction of new planners that rely on SMT solvers to deal with the most expressive extensions of PDDL.
Most notably, the most effective approaches for planning with PDDL\plus rely on compilations to SMT \cite{bryce2015a,cashmore2016}. 

\smtplan \cite{cashmore2016} appears to be the most effective of those planners.
It supports planning with continuous non-linear change and processes and the authors showed that those can be efficiently accounted for in an SMT representation.

Of all the planning benchmarks we tested it on, \smtplan was only able to solve the two simplest instances of the (non-temporal) \textit{blocks} domain.
Its performance on domain-independent temporal or non-temporal planning is thus largely below the ones of LCP and state-of-the-art planners.

\smtplan uses a state-oriented representation where each state is associated to a given happening that alters the state (corresponding to the start or end of an action).
Each state is represented by a set of boolean variables.
To each happening is associated the choice of a primitive action that implies constraints on the happening's state variables.

To understand the implication of this representation, we compare the encoding of both LCP and \smtplan on two problems of the \textit{rovers} domains. 
The two problems mainly differ by the number of objects in the domain, with a direct impact on the number of primitive actions.
The first instance (p01) has only 63 primitive actions while the second instance considered (p10) has 382.
Note that these numbers are low compared to standard planning benchmarks (e.g. the most difficult problem of the \textit{rovers} domain has 32,437 reachable primitive actions).

The number of disjunctive constraints (i.e. clauses in the SMT formula) as well as the number of variables is given in \autoref{tab:smt-size} for both LCP and \smtplan.
It can be seen that the size of the SMT formula at a given depth in \smtplan is directly impacted by the number of primitive actions, making it quickly untractable even on problems of modest size. The size of the encoding grows linearly with the number of happenings (depth).
On the other hand, LCP is mostly unaffected by the growth in problem size thanks to its use of partially instantiated actions.
On the down side, the encoding of LCP grows quadratically with the depth. Note that, in the case of LCP, a depth of 4 means that each action template might be duplicated 4 times (in the  \textit{rovers} domain, which has 9 action templates, the plan at depth 4 might contain up to 36 actions).
On the other hand, a durative action requires two happenings in \smtplan's representation, i.e., a depth of 4 would only allow two sequenced actions.

\begin{table}[htb!]
    \centering
    \setlength{\tabcolsep}{10pt}
    \renewcommand{\arraystretch}{1.3}
    \begin{tabular}{|c|cc|cc|}\hline
         & \multicolumn{2}{c|}{Rovers p01} & \multicolumn{2}{c|}{Rovers p10} \\
        Depth     & LCP & \smtplan & LCP & \smtplan \\ \hline
        1        & 109 / 115 &  1\,987 / 332 & 165 / 132 & 28\,760 / 1\,828 \\
        2        & 295 / 218   & 3\,987 / 664 & 399 / 235 & 59\,293 / 3\,656\\
        3        & 561 / 321 &  6\,161 / 996 & 713 / 338 &  89\,196 / 5\,484\\
        4         & 907 / 424  & 8\,248 / 1\,328 & 1\,107 / 441 & 119\,099 / 7\,312\\ \hline
    \end{tabular}\\[5pt]
    \caption{
        Number of disjunctive constraints (left) and variables (right) in the SMT formulas of LCP and \smtplan at various depths.
        Depth is the number $k$ of duplicated actions in LCP and the number of happenings in \smtplan.
    }
    \label{tab:smt-size}
\end{table}       

\section{Related Work}
\label{sec:related-work}

\subsubsection{Encodings for Graph-Plan}
A historical application of CP solvers to planning has been based on the graph-plan framework \cite{blum1997}.
Such planners build a synthetic data structure that captures causal relationships between primitive actions up to a given plan length.
While building such a data structure can be done in polynomial time, extracting a solution (or proving its absence) is combinatorial.
CP solvers have been leveraged to perform the plan extraction step in planners such as GP-CSP \cite{do2000} and CSP-PLAN \cite{lopez2003}.
Both planners use a grounded state-based encoding where the value of variables in each state is related to the previous state and primitive actions through a set of constraints.
Despite work on improving the encoding of the CSP with table constraints \cite{bartak2008}, the performance of such planners has been superseded by forward search planners.

\subsubsection{Plan-Space and Timeline-based Planners} 
Another line of research is the work on timeline-based planners \cite{cesta2009,chien2000a,frank2003} and lifted plan-space planners \cite{bit-monnot2016,ghallab1994}.
Both kind of planners use a time-oriented representation, and the chronicle model that we use originated in lifted plan-space planners \cite{ghallab1994}.
Those planners search in the space of partial plans where the current partial plan is extended with new actions during search.
Key aspects of the partial plan are represented in a CSP, including the actions' parameters and timepoints.

One difficulty is that the CSP is constructed online -- both constraints and variables are added to the CSP -- limiting the ability to use existing constraint solvers.
More problematic is that the set of effects that can be used to support a given condition is not fixed \emph{a priori} as new actions can be inserted in the partial plan.
As a result, the equivalent of our \textit{supported-by} constraints are often implemented in an \emph{ad hoc} way, outside of the main constraint engine.

The internal representation of those planners is however closely related to the encoding proposed in this paper.
One important difference is that plan-space planners use \emph{threats} to encode the consistency between support decisions.
We avoid the explicit handling of threats by the introduction of a new timepoint in effect tokens to represent the minimal persistence of an effect which is accounted for directly in the coherence constraint.
This is important to limit the size of the CSP, as an explicit encoding of threats would be cubic in the number of possible actions.

The proximity with those planners might make it possible to adapt the relaxations, heuristics and propagators developed into global constraints to our setting, such as the one tailored to reason on causal relationships in EUROPA and FAPE \cite{bernardini2007,bit-monnot-2016-ecai}.

\subsubsection{SMT-based Planners}
The development of SMT encodings for task planning in the last years has been motivated by the need to support richer planning problems, notably involving continuous change on numeric state variables in planners such as dReach \cite{bryce2015a} and \smtplan \cite{cashmore2016}.
As highlighted in \autoref{sec:comparison-smtplan}, it is our feeling that the support of more complex problems has been done at the detriment of those planners scalability.
Our objective with this paper is precisely to propose an alternative encoding that can \emph{(i)} be used to efficiently represent planning problems and \emph{(ii)} does not prevent its extension to problems with continuous change.
Our encoding already supports continuous state variables and -- unlike dReach and {\smtplan} -- action parameters with continuous domains.
As for \smtplan and dReach, our use of an SMT solver with non-linear arithmetic theories opens up the possibility of reasoning with non-linear continuous changes and processes.

\subsubsection{Constraint-based Scheduling} 
The consideration of optional activities in constraint solvers slowly narrows the gap that existed between planning and scheduling.
Our formulation of a bounded planning problem is indeed very close to a scheduling problem with optional activities over given temporal intervals, in the spirit of those in CP Optimizer \cite{laborie2008}.

Achieving good, domain-independent, performance in a such a setting would certainly require specialized global constraints.
This is a promising direction to tackle the current quadratic growth in the number of constraints of our representation.

Our coherence constraints are closely related to scheduling optional activities on unary resources and state-functions for which existing techniques could be easily adapted \cite{laborie2008,laborie2009,vilim2005a}.
The main difference here being the implicit choice of the state variable that is governed by several variables.
Support constraints are more challenging and to our knowledge have no straightforward mapping to global constraints in scheduling. 
Indeed, the closest equivalent (state-functions in CP Optimizer) lack the notions of conditions and effects.
While plan-space planners do provide propagators for such support constraints \cite{bit-monnot-2016-ecai,vidal2006}, those are highly tailored to their search mechanism and adapting them would require more work.

\section{Conclusion}

In this paper, we presented a constraint-based encoding for temporal planning.
The encoding is focused on handling the conditions and effects appearing in typical planning problems.
It leverages a fully lifted representation and allows the addition of arbitrary constraints.
These characteristics make it a good fit for a usage at the core of more general constraint-based planners or for embedding planning subproblems in CSPs.

Its usage in a simple planner, using an off-the-shelf SMT solver, shows improved performance with respect to state of the art temporal planners on planning problems with deadlines and time-windows.
The resulting planner largely outperforms existing SMT-based temporal planners on planning benchmarks.

One important contribution is the identification the small discrepancies that remain between the available global constraints in constraint-based scheduling and the requirements for efficiently handling task planning problems; going one step closer to closing the long standing gap between planning and scheduling.

\bibliographystyle{splncs04}
\bibliography{local} 

\end{document}